\documentclass{article}

\usepackage{arxiv}

\usepackage[utf8]{inputenc} 
\usepackage[T1]{fontenc}    
\usepackage{hyperref}       
\usepackage{url}            
\usepackage{booktabs}       
\usepackage{amsfonts}       
\usepackage{nicefrac}       
\usepackage{microtype}      
\usepackage{lipsum}		
\usepackage{graphicx}
\usepackage{natbib}
\usepackage{doi}

\usepackage[figuresright]{rotating}
\usepackage[caption=false, font=footnotesize]{subfig}

\usepackage{amssymb}
\usepackage{amsthm}
\usepackage{amsmath}
\usepackage[scientific-notation=true]{siunitx} 
\usepackage[normalem]{ulem}   

\usepackage[ruled,linesnumbered]{algorithm2e}
\usepackage[utf8]{inputenc} 
\usepackage[T1]{fontenc}    
\usepackage{hyperref}       
\usepackage{url}            
\usepackage{amsfonts}       
\usepackage{amsmath}
\usepackage{amssymb}
\usepackage{amsbsy}
\usepackage{lipsum}
\usepackage{graphicx}
\usepackage{float}
\usepackage[caption = false]{subfig}
\usepackage[export]{adjustbox}
\usepackage[section]{placeins}
\usepackage{layouts}
\usepackage[super]{nth}

\SetArgSty{textnormal}
\SetKwInput{KwInput}{Input}
\SetKw{Break}{break}
\newcommand{\var}{\texttt}

\usepackage{ulem,lipsum}
\newcommand\xoutpars[1]{\let\helpcmd\xout\parhelp#1\par\relax\relax}
\newcommand\soutpars[1]{\let\helpcmd\sout\parhelp#1\par\relax\relax}
\long\def\parhelp#1\par#2\relax{%
	\helpcmd{#1}\ifx\relax#2\else\par\parhelp#2\relax\fi%
}

\usepackage{amsfonts}       
\usepackage{amssymb}        
\usepackage{amsthm}         
\usepackage{amsmath}

\usepackage[ruled,linesnumbered]{algorithm2e}
\usepackage{graphicx}
\usepackage{float}
\usepackage[caption = false]{subfig}
\usepackage[super]{nth}
\usepackage{siunitx}        
\usepackage{tabularx}

\SetArgSty{textnormal}
\SetKwInput{KwInput}{Input}
\SetKw{Break}{break}

\newcolumntype{C}{>{\centering\arraybackslash}X}
\newcolumntype{R}{>{\raggedleft\arraybackslash}X}
\newcolumntype{L}{>{\raggedright\arraybackslash}X}

\usepackage{booktabs}
\usepackage{multirow}
\usepackage{complexity}

\usepackage{array, longtable, tabularx}
\usepackage{pdflscape}

\usepackage{graphicx}
\usepackage[super]{nth}

\usepackage{xcolor}

\usepackage{lineno,hyperref}
\modulolinenumbers[5]

\title{Learning to extrapolate using continued fractions:
	Predicting the critical temperature of superconductor materials}

\author{ \href{https://orcid.org/ 0000-0003-2570-5966}{\includegraphics[scale=0.06]{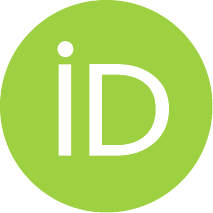}\hspace{1mm}Pablo Moscato}\\ 
	School of Information and Physical Sciences\\
	The University of Newcastle\\
	University Drive, Callaghan, NSW 2308, Australia \\
	\texttt{Pablo.Moscato@newcastle.edu.au} \\
	\texttt{https://www.newcastle.edu.au/profile/pablo-moscato}\\
	\And
	\href{https://orcid.org/0000-0002-0598-0867}{\includegraphics[scale=0.06]{orcid.pdf}\hspace{1mm}Mohammad N.~Haque}\\
	ResTech Pty Ltd,\\
	CE Building, Design Drive, Callaghan, NSW 2308, Australia\\
	\texttt{Mohammad.Haque@restech.net.au} \\
	\And
	\href{https://orcid.org/0000-0002-2222-7519}{\includegraphics[scale=0.06]{orcid.pdf}\hspace{1mm} Kevin Hunag} \\
	University of Washington\\
	Bill \& Melinda Gates Center, 3800 E Stevens Way NE Seattle, WA 98195 USA\\
	\texttt{kehuang@cs.washington.edu} \\
	\And
	\href{https://orcid.org/0000-0003-0200-063X}{\includegraphics[scale=0.06]{orcid.pdf}\hspace{1mm} Julia Sloan} \\
	California Institute of Technology\\
	1200 E California Blvd, Pasadena, CA 91125, USA\\
	\texttt{jsolan@caltech.edu} \\
	\And
	\href{https://orcid.org/0000-0002-2222-7519}{\includegraphics[scale=0.06]{orcid.pdf}\hspace{1mm} Jonathon Corrales de Oliveira} \\
	7668 SW $152^{nd}$ Ave Apt 208, Miami, Fl 33193, USA\\
	\texttt{jcjonathonc5@gmail.com} \\
}

\renewcommand{\shorttitle}{\textit{arXiv} Template}

\hypersetup{
	pdftitle={A template for the arxiv style},
	pdfsubject={q-bio.NC, q-bio.QM},
	pdfauthor={David S.~Hippocampus, Elias D.~Striatum},
	pdfkeywords={First keyword, Second keyword, More},
}

\begin{document}
	\maketitle
	
	\begin{abstract}
		In the field of Artificial Intelligence (AI) and Machine Learning (ML), the approximation of unknown target functions $y=f(\mathbf{x})$ using limited instances $S={(\mathbf{x^{(i)}},y^{(i)})}$, where $\mathbf{x^{(i)}} \in D$ and $D$ represents the domain of interest, is a common objective. We refer to $S$ as the training set and aim to identify a low-complexity mathematical model that can effectively approximate this target function for new instances $\mathbf{x}$. Consequently, the model's generalization ability is evaluated on a separate set $T=\{\mathbf{x^{(j)}}\} \subset D$, where $T \neq S$, frequently with $T \cap S = \emptyset$, to assess its performance beyond the training set.\\
		However, certain applications require accurate approximation not only within the original domain $D$ but also in an extended domain $D'$ that encompasses $D$. This becomes particularly relevant in scenarios involving the design of new structures, where minimizing errors in approximations is crucial. For example, when developing new materials through data-driven approaches, the AI/ML system can provide valuable insights to guide the design process by serving as a surrogate function. Consequently, the learned model can be employed to facilitate the design of new laboratory experiments.\\
		In this paper, we propose a method for multivariate regression based on iterative fitting of a continued fraction, incorporating additive spline models. We compare the performance of our method with established techniques, including \textit{AdaBoost}, \textit{Kernel Ridge}, \textit{Linear Regression}, \textit{Lasso Lars}, \textit{Linear Support Vector Regression}, \textit{Multi-Layer Perceptrons}, \textit{Random Forests}, \textit{Stochastic Gradient Descent}, and \textit{XGBoost}. To evaluate these methods, we focus on an important problem in the field: predicting the critical temperature of superconductors based on physical-chemical characteristics.
	\end{abstract}

	\keywords{Regression \and Continued Fractions \and Superconducting materials \and Superconductivity}

	\section{Introduction}
	
	Superconductors are remarkable materials that exhibit the extraordinary property of conducting electrical current with zero resistance. This unique characteristic has led to a wide range of applications, with Magnetic Resonance Imaging (MRI) systems being used globally as a crucial medical tool for producing detailed images of internal organs and tissues. Additionally, in the face of increasing energy demands driven by renewable energy sources and innovations like solar cars, superconductors hold the potential for efficient energy transfer.
	
	The elimination of electrical resistance in superconductors significantly reduces energy wastage during current transmission from one location to another. However, a major limitation of existing superconductors is their reliance on extremely low temperatures, known as critical temperatures ($T_c$), to achieve zero resistance. Typically, these critical temperatures are incredibly cold, often around -196\textdegree C, and vary depending on the specific superconducting material~\cite{Superconductivity:2018-Hamidieh-DataDriven}. Predicting the critical temperature ($T_c$) of superconductors has therefore become a topic of great interest in the field of materials science.
	
	In this study, we leverage various machine learning techniques and propose a novel approach based on multivariate continued fractions to develop mathematical models capable of predicting the critical temperature of superconductors. Our models rely solely on the characterization of the chemical structure of the superconducting material, uncovering hidden information within. Accurate prediction of $T_c$ for superconductors will greatly enhance our ability to harness their potential, ushering in a new era of possibilities in multiple fields.

	\subsection{Continued Fraction Regression}

	In 2019, a new approach for multivariate regression using continued fractions was introduced in~\cite{DBLP:conf/cec/SunM19} and compared with a state of the art genetic programming method for regression. 
	A year later, this technique's results on 354 datasets from the physico-chemical sciences were presented in 
	\cite{DBLP:conf/cec/MoscatoSH20} and compared with some of the state-of-the-art top 10 regression techniques. The new method was the top-ranked performer in the training set in 352 out of the 354, and it was the also first in terms of generalisation in 192, more than half of the total of times of all other 10 methods combined. The figure of merit was the Mean Squared Error.  
	
	We named this known approach as `Continued Fraction Regression', or CFR. The best existing algorithm currently utilizes a memetic algorithm for optimizing the coefficients of a model that approximates a target function as the convergent of a continued fraction~\cite{DBLP:conf/cec/SunM19,DBLP:conf/cec/MoscatoSH20,moscato2019analytic}.
	Memetic Algorithms are well-established research areas in the field of Evolutionary Computation and the IEEE had established a Task Force in Computational Intelligence for their study. Therefore, it is important to refer the readers to some of the latest references and reviews on the field
	~\cite{DBLP:reference/crc/CottaM07,DBLP:series/sci/Moscato12,DBLP:reference/sp/CottaMM18,Moscato2019:MA-accelerated-intro,DBLP:books/sp/19/MoscatoM19}. 
	Very recently, continued fraction regression has been used 
	to obtain analytical approximations of the minimum electrostatic energy configuration of $n$ electrons, when the charges are constrained to be on the surface sphere, i.e. the celebrated Thomson Problem~\cite{Moscato2023}.

	Some basic introduction on analytic continued fraction approximation is perhaps necessary. A continued fraction for a real value $\alpha$ is of the following form 
	\eqref{general-continued-continued-fraction-expansion} and may be finite or infinite~\cite{sun}, according to $\alpha$ being a rational number or not, respectively.
	\begin{equation}
		\alpha= a_0 + \cfrac{b_1}{a_1 + \cfrac{b_2}{a_2 + \ldots}}
		\label{general-continued-continued-fraction-expansion}
	\end{equation}
	Euler's proved a mathematical formula that allows us to write a sum of products as a continued fraction\eqref{euler-continued-fraction-formula}:
	\begin{multline}
		\beta = a_0 + a_0a_1 + a_0a_1a_2 + \ldots + a_0a_1a_2\dots a_n \\
		= \cfrac{a_0}{1 - \cfrac{a_1}{1 + a_1 - \cfrac{a_2}{1 + a_2 - \cfrac{\ddots}{\ddots \frac{a_{n-1}}{1 + a_{n-1} - \frac{a_n}{1+a_n}}}}}}.
		\label{euler-continued-fraction-formula}
	\end{multline}
	
	This simple yet powerful equation reveals how infinite series can be written as infinite continued fractions, meaning that continued fractions can be a good general technique to approximate analytic functions thanks to the improved optimization methods such as those provided by memetic algorithms~\cite{moscato2019analytic}. Indeed, CFR has already demonstrated to be an effective regression technique on the real-world benchmark provided by the \textit{Penn Machine Learning Database}~\cite{moscato2019analytic}. 
	
	In this paper, we will use Carl Friedrich Gauss' mathematical notation for generalized continued fractions \cite{knotation} (i.e. a compact notation where ``K'' stands for the German word ``Kettenbruch'' which means `Continued Fraction'). Using this notation, we may write the continued fraction in \eqref{general-continued-continued-fraction-expansion} as:
	
	\begin{equation}
		\alpha = a_0 + \K_{i = 1}^{\infty} \frac{b_i}{a_i},
	\end{equation}
	\noindent
	thus the problem of finding an approximation of an unknown target function of $n$ variables $\mathbf{x}$ given a training dataset of $m$ samples 
	$S=\{ ( \mathbf{x^{ (i) }}, y^{(i)} )\}$ 
	is that of finding the set of functions 
	$F=\{a_0(x)..., b_1(x), ...\}$
	such that a certain objective function is minimized; i.e. we aim to find
	\begin{equation}
		f(\mathbf{x}) = a_0(\mathbf{x}) + \K_{i = 1}^{\infty} \frac{b_i(\mathbf{x})}{a_i(\mathbf{x})}. 
	\end{equation}

	\section{Materials and Methods}

	
	

	\subsection{A new approach: Continued Fractions with Splines}
	
	In previous contributions~\cite{DBLP:conf/cec/SunM19,DBLP:conf/cec/MoscatoSH20,moscato2019analytic}, a memetic algorithm was always employed to find the approximations. Here, we present another method to fit continued fraction representations by iteratively fitting splines. 
	
	Splines provide a regression technique that involves fitting piecewise polynomial functions to the given data~\cite{de1978Spline}. The domain is partitioned into intervals at locations known as ``knots''. Then, a polynomial model of degree $n$ is separately fitted for each interval, generally enforcing boundary conditions including continuity of the function as well as the continuity of the first $(n\text{-}1)$-order derivatives at each of the knots. Splines can be represented as a linear combination of basis functions, of which the standard is the B-spline basis. Thus, fitting a spline model is equivalent to fitting a linear model of basis functions. We refer to Hastie {\it et al}. \cite{principles} for the particular definition of the B-spline basis.
	
	First, when all the functions $b_i(\mathbf{x})=1$, for all $i$, we have a \textit{simple continued fraction} representation, and we can write it as:
	
	\begin{equation}
		f(\mathbf{x}) = g_0(\mathbf{x}) + \cfrac{1}{g_1(\mathbf{x}) + \cfrac{1}{g_2(\mathbf{x}) + \cfrac{1}{g_3(\mathbf{x}) + ...}}}.
	\end{equation}
	\noindent
	Note that for a term $g_i(\mathbf{x})$, we say that it is at ''depth'' $i$.
	
	Finding the best values for the coefficients in the set of functions $\{g_i(\mathbf{x})\}$, can be addressed as a non-linear optimization problem as in 
	\cite{DBLP:conf/cec/SunM19,DBLP:conf/cec/MoscatoSH20,moscato2019analytic}. However, despite the great performance of that approach, we aim to introduce a faster variant that can scale well to larger datasets such as this one. 
	
	Towards that end, and thinking about the scalability, we fit the model iteratively by depth as follows: we first consider only the first term, $g_0(\mathbf{x})$ (at depth 0), ignoring all other terms. We fit a model for the first term using predictors $\mathbf{x}$ and the target $f(x)$. Next, we consider only the first and second depths, with the terms $g_0(\mathbf{x})$ and $g_1(\mathbf{x})$, ignoring the rest. We then fit $g_1(\mathbf{x})$ using the previously fit model for $g_0(\mathbf{x})$. For example, truncating the expansion at depth 1, we have that 
	\begin{equation}
		g_1(\mathbf{x}) = \frac{1}{f(\mathbf{x}) - g_0(\mathbf{x})}.
	\end{equation}
	
	Thus, we fit $g_1(\mathbf{x})$ using the predictors $\mathbf{x}$ and the target $(f(\mathbf{x}) - g_0(\mathbf{x}))^{-1}$. We label this target  as $y^{(1)}$. We repeat this process, fitting a new model by truncating at the next depth by using the models fit from previous depths and iterations. 
	
	We have that at depth $i > 0$, the target $y^{(i)}$ for the model $g_i(\mathbf{x})$ is $(\epsilon_{i - 1})^{-1}(\mathbf{x})$, where $\epsilon_{i - 1}(\mathbf{x})$ is the residual of the previous depth's model, $y^{(i-1)} - g_{i - 1}(\mathbf{x})$. 
	
	One notable characteristic of this approach is that if any model $g_i(\mathbf{x})$, $i > 0$ evaluates to 0, then we will have a pole in the continued fraction, which is often spurious. To remedy this, we modify the structure of the fraction such that each fitted $g_i(\mathbf{x})$, $i > 0$ is encouraged to be strictly positive on the domain of the training data. To do this, we add a constant $C_i$ to $\epsilon_i$ when calculating the target $y^{(i + 1)}$, where $C_i = |\min_x \epsilon_i|$. Thus, the targets $y^{(i)}$ for $i > 0$ are all non-negative, encouraging each $g_i(\mathbf{x})$, $i> 0$, to be strictly positive. For example, for $g_1(\mathbf{x})$, we would have that the target $y^{(1)} = (f(\mathbf{x}) - g_0(\mathbf{x}) + C_1)^{-1}$. Of course, we must then subtract $C_i$ from $g_{i - 1}(\mathbf{x})$ in the final continued fraction model.
	
	We have found that data normalization often results in a better fit using this approach. It is sufficient to simply divide the targets uniformly by a constant when training and multiply by the same constant for prediction. We denote this constant parameter $\texttt{norm}$.
	
	A good choice of the regression model for each $g_i(\mathbf{x})$ is a spline since they are well-established. For reasons stated in the next section, the exception is the first term $g_0(\mathbf{x})$, which is a linear model. We use an additive model to work with multivariate data where each term is a spline along a dimension. That is, given $m$ predictor variables, we have that 
	
	\begin{equation} \label{eq:spline_model}
		g_i(\mathbf{x}) = \sum_{j=1}^{m} f_j(x_j)
	\end{equation}
	\noindent
	for each term $g_i(\mathbf{x})$, $i > 0$, where each function $f_j$ is a cubic spline along variable $j$. That is, $f_j$ is a piecewise polynomial of degree 3 and is a function of variable $j$. 
	
	We implement the splines with a penalized cubic B-spline basis. That is, $f_j(\mathbf{x}) = \sum_{i=1}^{k} \beta_k B_k(x_j)$, where each $B_i(x)$ is one of $k$ cubic B-spline basis functions along dimension $j$ and corresponds to one of $k$ knots. We use the following loss function $L\left(\mathbf{B}\left(\mathbf{x}, \mathbf{y}, \pmb{\beta}\right)\right)$, i.e. 
	\begin{equation}
		L\left(\mathbf{B}\left(\mathbf{x}, \mathbf{y}, \pmb{\beta}\right)\right) =   \Vert \mathbf{y} - \mathbf{B} \pmb{\beta} \Vert^2 + \lambda \sum_{j=0}^{m} \pmb{\beta}^T \mathbf{P_j} \pmb{\beta}
	\end{equation}
	\noindent
	where $\mathbf{B}$ is the matrix of cubic B-spline basis functions for all variables, $\pmb{\beta}$ is the vector of all of the weights, and $\mathbf{P_j}$ is the associated second derivative smoothing penalty matrix for the basis for the spline $f_j$. This is standard for spline models \cite{principles}. The pseudocode for this approach is shown in Algorithm~\ref{algo:spline-cfr}.
	
	\begin{algorithm}
		\SetAlgoLined 
		\KwInput{Training data $\mathcal{D} = \{(\mathbf{x_1}, f(\mathbf{x_1}), ..., (\mathbf{x_n}, f(\mathbf{x_n}))\}$ and parameters $\lambda$, $k$, $\var{norm}$, and $\var{max\_depth}$}
		\tcc{Let $n$ be the number of samples; $m$ be the number of variables}
		\tcc{$\mathbf{X} \in \mathbb{R}^{n \times m}$ be data matrix and $\mathbf{y} \in \mathbb{R}^{n}$ be the vector of targets.}
		$\texttt{knot\_indices} = \{\}$
		
		$\mathbf{y^{(0)}} \leftarrow \mathbf{y} / \var{norm}$
		
		\For{$i \leftarrow$ 0, 1, ..., \var{max\_depth}}{
				\eIf{$i = 0$} {
					\tcc{$g_0$ is a linear model parameterized by $\mathbf{\beta}$, and is fit with least squares.}
					
					$\mathbf{\beta} \leftarrow \text{argmin}_{\mathbf{\beta}} \displaystyle \Vert \mathbf{y^{(0)}} - \mathbf{X}\mathbf{\beta} \Vert^2$
				}
					{
						\tcc{$g_i$ be an additive spline model as given in equation \eqref{eq:spline_model}, parameterized by $\mathbf{\beta}$. For each predictor variable, the knots are at the samples indexed by the first $k$ indices in $\texttt{knot\_indices}$}
						
						\For {$j \leftarrow$ 1, 2, ..., $m$} {
							$f_j \leftarrow$ new SplineModel()
							
							\For {each index $p$ in $\texttt{knot\_indices}$} {
								$f_j \leftarrow$ AssignKnotAt($\mathbf{X}[p][j]$)
							}
						}
						$g_i = \sum_{j=1}^{m} f_j(x_j)$
						
						\tcc{Construct the splines, and fit with regularized least squares}
						
						$\mathbf{B} \leftarrow$ BSplineBasisMatrix($g_i$.knots)
						
						$\mathbf{P}_j \leftarrow$ BSplinePenaltyMatrix($f_j$: for each $f_j$ in $g_i$)
						
						$\mathbf{\beta} \leftarrow \text{argmin}_{\mathbf{\beta}} \displaystyle \Vert \mathbf{y^{(i)}} - \mathbf{B}\mathbf{\beta} \Vert^2 + \lambda \sum_{j = 1}^{m} \mathbf{\beta}^T\mathbf{P}_j \mathbf{\beta}$
					}

					\tcc{Compute $\mathbf{\epsilon_i}$, the vector of residuals of the $i$th model, and then compute the targets and knot locations for the next depth.}
					$\mathbf{\epsilon}_i \leftarrow \mathbf{y^{(i)}} - g_i(\mathbf{X})$
					
					$C_i \leftarrow | \min_x \mathbf{\epsilon}_i |$
					
					$\mathbf{y^{(i + 1)}} \leftarrow (\mathbf{\epsilon_{i}} + C_{i})^{-1}$
					
					$\texttt{knot\_indices} \leftarrow$ SelectKnots($\mathbf{\epsilon_i}$) 
					
				}
				
				The estimate for $f(\mathbf{\mathbf{x}})$ at $max\_depth$ is:
				\begin{equation*}
					\approx \var{norm} \; \cdot \left[ g_0(\mathbf{x}) - C_0 + \K_{i = 1}^{\var{max\_depth}} \frac{1}{g_i(\mathbf{x}) - C_i} \right]
				\end{equation*}
				\caption{Iterative CFR using additive spline models with adaptive knot selection}\label{algo:spline-cfr}
			\end{algorithm}

			\subsection{Adaptive knot selection}

			\begin{figure}[H]
				\centering
				\subfloat[depth 3]{\includegraphics[width =7.0cm]{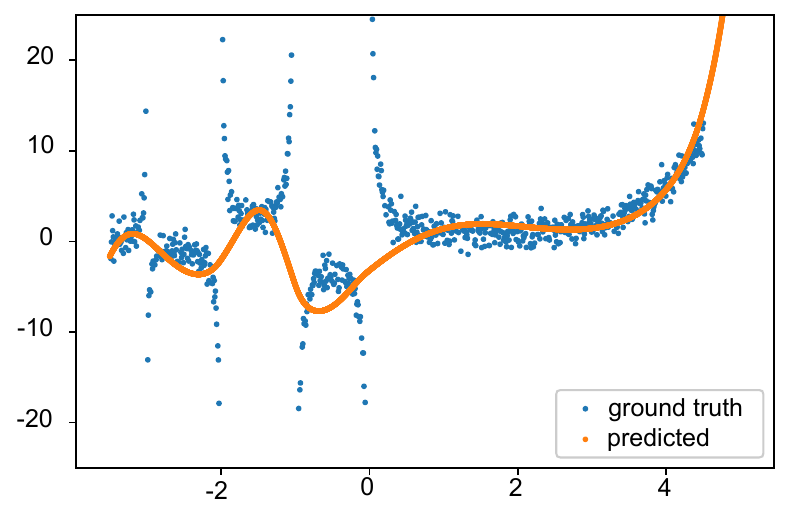}} 
				\subfloat[depth 5]{\includegraphics[width =7.0cm]{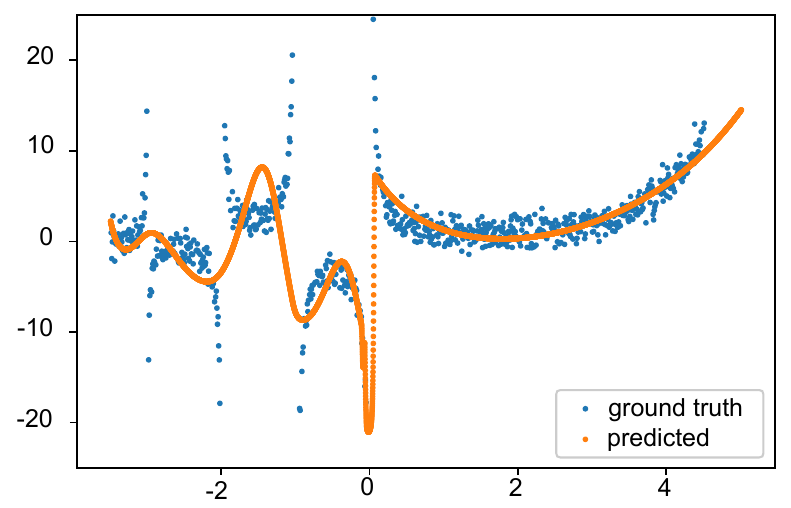}} \\
				\subfloat[depth 10]{\includegraphics[width =7.0cm]{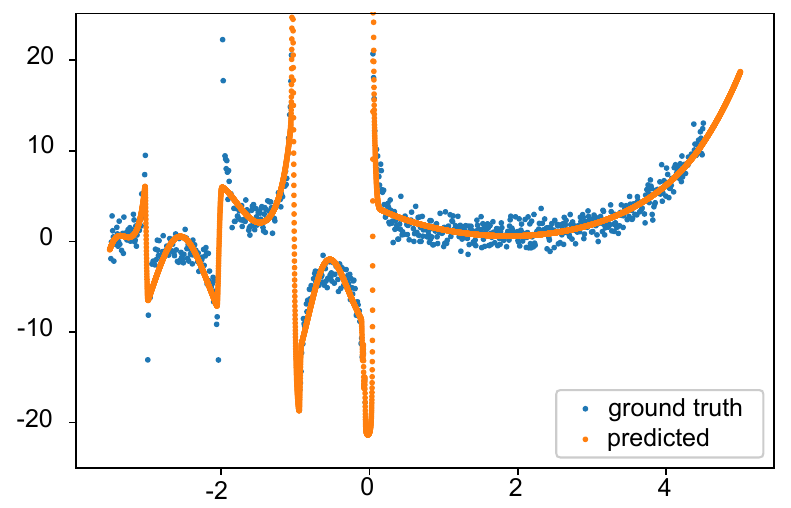}} 
				\subfloat[depth 15]{\includegraphics[width =7.0cm]{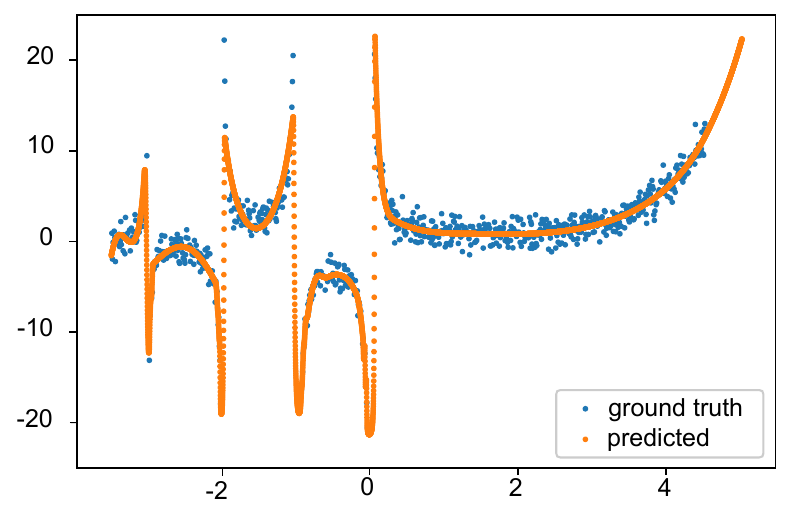}} 
				
				\caption{Examples of the fit obtained by the \textit{Spline Continued Fraction} using a dataset generated thanks to the gamma function with added noise. We present several continued fractions with depths of 3 (a), 5 (b), 10 (c), and 15 (d). In this example, the number of knots $k$ was chosen to be 3, $norm$ = 1, and $\lambda$ = 0.1.\label{fig:gamma}}
			\end{figure}

			The iterative method of fitting continued fractions also allows for an adaptive method of selecting knot placements for the additive spline models. For the spline model $g_i(\mathbf{x})$ at depth $i > 0$, we use all of the knots of the spline model $g_{i - 1}(\mathbf{x})$ at depth $i - 1$. Then, for each variable, we place $k$ new knots at the unique locations of the $k$ samples with the highest absolute error from the model $g_{i - 1}(\mathbf{x})$ at depth $i - 1$. As the points with the highest error can be likely to be very close to each other, we impose the condition that we take the samples with the highest error, but they must have alternating signs.
			
			That is, for $g_i(\mathbf{x})$, $i > 0$, we select $k$ knots, with the first knot at the location of the sample with the highest absolute error computed from the model $g_{i - 1}(\mathbf{x})$. For the rest of the knots, the $j^\text{th}$ knot is selected at the sample's location with the next highest absolute error than the sample used for the $(j - 1)^\text{th}$ knot. Nevertheless, only if the sign of the (non-absolute) error of that sample is different from the sign of the (non-absolute) error of the sample used for the $(j - 1)^\text{th}$ knot. Otherwise, we move on to the next highest absolute error sample, and so on, until we fulfill this condition. This knot selection procedure is shown in Algorithm~\ref{algo:adaptive-knot}. Note that we let $g_0$ be a linear model as there is no previous model to obtain the knot locations from.

			\begin{algorithm}
				\SetAlgoLined
				\KwInput{$\epsilon_i$}
				\tcc{Given the vector of residuals $\mathbf{\epsilon_i}$ of the spline model at depth $i$, select the knot placements for the next spline model at depth $i + 1$}
				
				\tcc {Sort by indices of highest absolute error}
				
				$\texttt{abs\_error} \leftarrow$ elementWiseAbsoluteValue($\mathbf{\epsilon_i}$)
				
				$\texttt{highest\_error\_indices} \leftarrow$ argsortDecreasing($\texttt{abs\_error}$)
				
				\tcc {Take the top $k$ highest order indices, such that each error term has opposite sign of the last}
				
				$\texttt{current\_sign} \leftarrow null$
				
				$\texttt{knots\_added} \leftarrow 0$
				
				\For {each $\texttt{i}$ in $\texttt{highest\_error\_indices}$} {
					\If {$\texttt{knots\_added} \geq k$} {
						\Break
					}
					\If {sign($\epsilon_i$[$\texttt{i}$] $\neq \texttt{current\_sign}$)} {
						
						$\texttt{current\_sign} \leftarrow$ sign($\epsilon_i$)
						
						$\texttt{knot\_indices}$.append($\epsilon_i$[$\texttt{i}$])
						
						$\texttt{knots\_added} \leftarrow \texttt{knots\_added} + 1$
					}
					
				}
				\Return $\texttt{knot\_indices}$
				
				\caption{SelectKnots (Adaptive Knot Selection)}\label{algo:adaptive-knot}
			\end{algorithm}

			The goal of using additive spline models with the continued fraction is to take advantage of the continued fraction representation's demonstrated ability to approximate general functions (see the discussion on the relationship with Pad\'{e} approximants in~\cite{moscato2019analytic}). The fraction's hierarchical structure allows for the automatic introduction of variable interactions, which is not included in the additive models individually that constitute the fraction. The iterative approach to fitting allows for a better algorithm for knot selection.
			
			An example of this algorithm modeling the well-known gamma function (with standard normally distributed noise added) is demonstrated in Fig.~\ref{fig:gamma}. Here, we showed how the fitting to gamma is affected by different values of depths (3, 5, 10, 15) in \textit{Spline Continued Fraction}. As desired, it is evident from the figure that \textit{Spline Continued Fraction} with more depth fits better with the data.

			\subsection{Data and Methods in the Study}
			
			We used the superconductivity dataset, also used by
			Hamidieh \cite{Superconductivity:2018-Hamidieh-DataDriven}, from the UCI Machine Learning repository\footnote{\url{https://archive.ics.uci.edu/ml/datasets/Superconductivty+Data}}. The website contains two files. In this work, we have only used the
			\textit{train.csv} file, which contains information of 21263 superconductors along with the critical temperature and a total of 81 attributes for each of them.  
			
			We conducted two main studies to see the generalization capabilities of many regression algorithms. We denote them as the \textit{Out-of-Sample} and \textit{Out-of-Domain}, respectively.  For the \textit{Out-of-Sample} study, the data is randomly partitioned into 2/3rds training data and 1/3rd test data. Each model was fit on the training data, and the \textit{RMSE} is calculated on the separated test portion of the data. 
			
			For the \textit{Out-of-Domain} study, the data was partitioned such that the training samples are always extracted from the set of samples with the lowest 90\% of critical temperatures. For the test set, the samples come from the highest 10\% of critical temperatures. It turned out the lowest 90\% have critical temperatures < 89~K, whereas the highest 10\% have temperatures greater or equal to 89~K that range from 89~K to 185~K (we highlight that the range of variation of the test set is more than the one of the training set making the generalization a challenging task).
			For each of the 100 repeated runs of \textit{Out-of-Domain} test, we have randomly taken 1/2 of the training set (from lowest 90\% of the observed value) to train the models and the same ratio from the test data (from 10\% of the highest actual value) to estimate the model performance.
			This said the \textit{Out-of-Domain} study allows us to see the capacity of several regression models in ``predicting'' on a set of materials that have higher critical temperatures, meaning that generalization, in this case, is strictly connected with the extrapolation capacity of the fitted models.  We executed both the \textit{Out-of-Sample} and \textit{Out-of-Domain} tests for 100 times to help us validate our conclusions with statistical results. 
			
			The \textit{Spline Continued Fraction} model had a depth of 5, five knots per depth, a normalization constant of $1000$, and a regularization parameter $\lambda$ of 0.5. 
			These parameters resulted from a one-dimensional non-linear model fitting to problems like the gamma function with noise (already discussed in Fig.~\ref{fig:gamma}) and others such as fitting the function $f(x)=\sin(x)/x$. The parameters were selected empirically using these datasets, and no problem-specific tuning on the superconductivity datasets was conducted.
			
			The final model was then iteratively produced by beginning at a depth of 1 and increasing the depth by one until the error was greater than the one observed for a previous depth (which we considered as a proxy for overfitting the data). 
			
			To evaluate the performance of the \textit{Spline Continued Fraction} (\texttt{Spln-CFR}) introduced in this paper with other state-of-the-art regression methods, 
			{we used a set of 11 regressors from two popular Python libraries (\textit{XGBoost}~\cite{Chen:2016:XGBoost} and \textit{Scikit-learn} machine learning library~\cite{scikit-learn})}. 
			The name of the regression methods are listed as follows:
			\begin{itemize}
				\item \textit{AdaBoost} (\texttt{ada-b})
				\item \textit{Gradient Boosting} (\texttt{grad-b})
				\item \textit{Kernel Ridge} (\texttt{krnl-r})
				\item \textit{Lasso Lars} (\texttt{lasso-l})
				\item \textit{Linear Regression} (\texttt{l-regr})
				\item \textit{Linear SVR} (\texttt{l-svr})
				\item \textit{MLP Regressor} (\texttt{mlp})
				\item \textit{Random Forest} (\texttt{rf})
				\item \textit{Stochastic Gradient Descent} (\texttt{sgd-r}) 
				\item \textit{XGBoost} (\texttt{xg-b})
				
			\end{itemize}
			The \textit{XGBoost} code is available as an open-source package\footnote{https://github.com/dmlc/xgboost}. The parameters of the \textit{XGBoost} model were the same as used in Hamidieh (2018)~\cite{Superconductivity:2018-Hamidieh-DataDriven}. We kept the parameters of other machine learning algorithms the same as Scikit defaults.
			
			All executions of the experiments were performed on an Intel$^{\circledR{}}$ Core$^{\text{TM}}$ i7-9750H hex-core based computer with hyperthreading and 16GB of memory. The machine was running on Windows 10 operating system. We used Python v3.7 to implement the \textit{Spline Continued Fraction} using pyGAM~\cite{daniel_serven_2018_1476122} package. All experiments were executed under the same Python runtime and computing environment.

			\section{Results}

			\begin{table}[H] 
				\caption{Results form the 100 runs of the proposed {Spline Continued Fraction} and ten regression methods all trained on the dataset, with median of Root Mean Squared Error (\textit{RMSE}) and standard deviation as the uncertainty of error.\label{tab:res-all-models}}
				\begin{tabularx}{\textwidth}{CCC}
					\toprule
					\multirow{2}{*}{\textbf{Regressor}} & \multicolumn{2}{r}{\textbf{Median RMSE Score $\pm$ Std}}\\
					\cline{2-3}{} & {\textit{Out-of-Sample}} & {\textit{Out-of-Domain}}\\
					\midrule
					{Spln-CFR}	&	10.989 $\pm$ 0.382	&	\textbf{36.327 $\pm$  1.187}  \\
					{xg-b}	&	\textbf{9.474 $\pm$ 0.190}	&	37.264   $\pm$ 0.947 \\
					{rf}	&	9.670 $\pm$ 0.197	&	38.074 $\pm$ 0.751 \\
					{grad-b}	&	12.659 $\pm$ 0.178	&	39.609  $\pm$ 0.619 \\
					{l-regr}	&	17.618 $\pm$ 0.187	&	41.265 $\pm$ 0.466 \\
					{krnl-r}	&	17.635 $\pm$ 0.163	&	41.427 $\pm$ 0.464 \\
					{mlp}	&	19.797 $\pm$ 5.140	&	41.480  $\pm$ 9.640 \\
					{ada-b}	&	18.901  $\pm$ 0.686	&	47.502 $\pm$ 0.743 \\
					{l-svr}	&	26.065 $\pm$ 7.838	&	47.985  $\pm$ 1.734 \\
					{lasso-l}	&	34.234 $\pm$ 0.267	&	74.724 $\pm$ 0.376 \\
					{sgd-r}\textsuperscript{1}	&	N.R.	&	N.R. \\ 
					\bottomrule
				\end{tabularx}
				\noindent{\footnotesize{\textsuperscript{1} The \textit{Stochastic Gradient Descent} Regressor (\texttt{sgd-r}), without parameter estimation, predicted unreasonable high values and the predicted error measure is extreme. Hence, we are not reporting (N.R.) the performance of \texttt{sgd-r} and have omitted it from further analysis.}}
			\end{table}
			
			Table~\ref{tab:res-all-models} presents the results of the regression methods along-with with those of the \textit{Spline Continued Fraction} approach for both of \textit{Out-of-Sample} and \textit{Out-of-Domain} studies.
			The median \textit{RMSE} value obtained from 100 runs is taken as the \textit{Out-of-Sample RMSE} estimate.
			
			For each of the 100 repeated runs of \textit{Out-of-Domain} test, we estimate the model performance via the \textit{Out-of-Domain RMSE} score. The median \textit{RMSE} score obtained from this test performance is reported in Table~\ref{tab:res-all-models} as \textit{Out-of-Domain RMSE}. We also report on some other descriptive statistics like, for instance, the number of times that the regressor correctly predicted a material to have a critical temperature greater or equal to 89~K.

			\begin{figure}[H]
					\centering
					\subfloat[Heatmap for \textit{Out-of-Sample} test]{\includegraphics[width=6.5 cm]{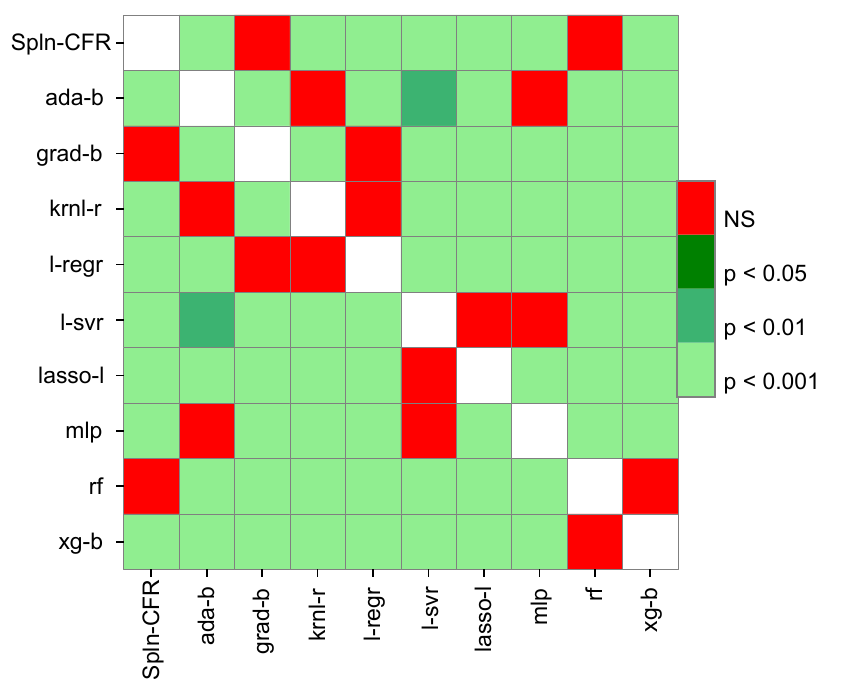}}\label{fig:heatmap-out-sample}
					\subfloat[Critical Diagram plot for \textit{Out-of-Sample} test]{\includegraphics[width=9.0 cm]{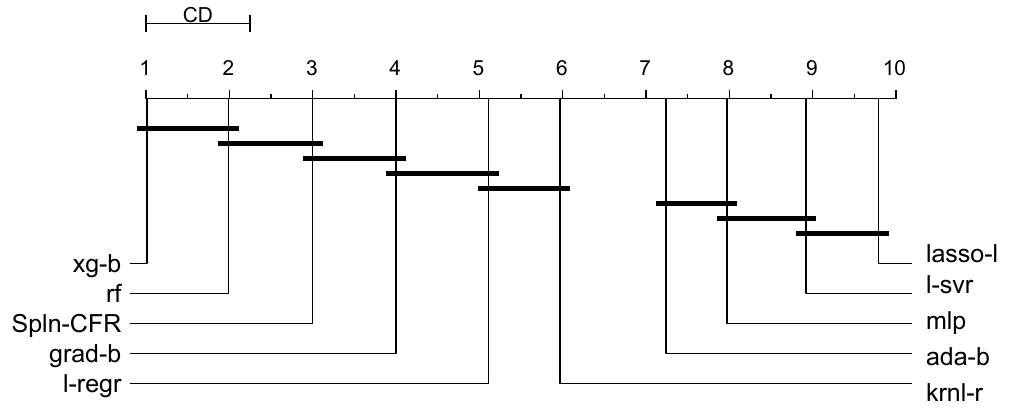} }\label{fig:cd-plot-out-sample}
				\caption{Statistical Comparison of the regressors for the \textit{Out-of-Sample} test. a) Heatmap showing the significance levels of $p$-values obtained by the Friedman Post-hoc Test and b) Critical difference (CD) plot showing the statistical significance of rankings achieved by the regression methods.\label{fig:stat-test-out-sample}}
			\end{figure}

			\begin{figure}[H]
					\centering
					\subfloat[Heatmap for \textit{Out-of-Domain} test]{\includegraphics[width=6.5 cm]{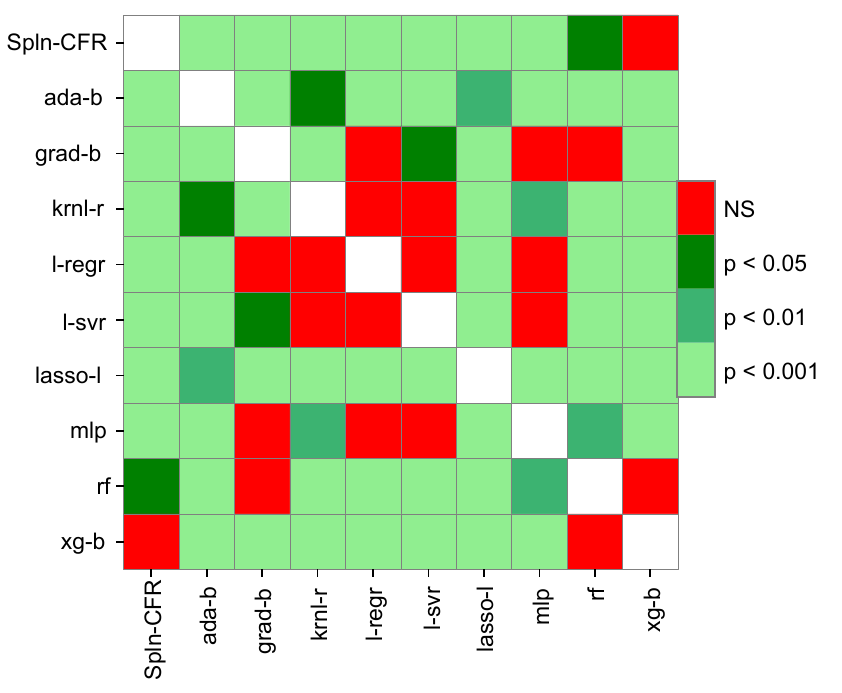}}\label{fig:heatmap-out-domain}
					\subfloat[Critical Diagram plot for \textit{Out-of-Domain} test]{\includegraphics[width=9.0 cm]{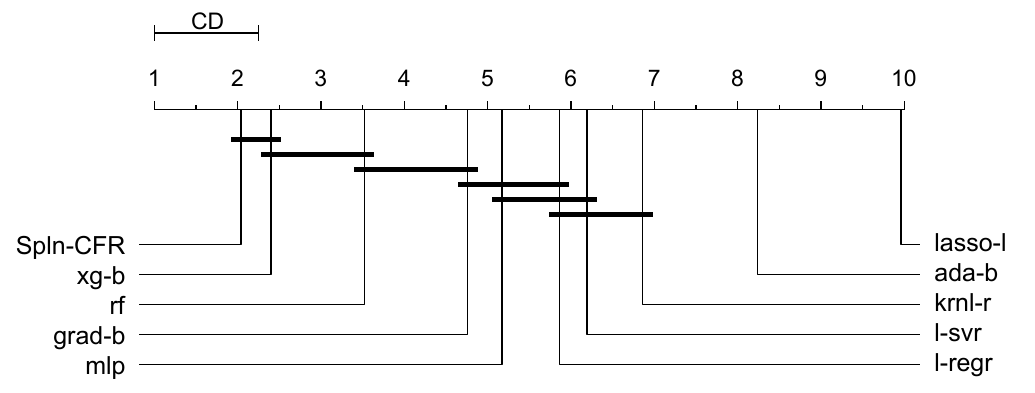}}\label{fig:cd-plot-out-domain}
				\caption{Statistical Comparison of the regressors for the \textit{Out-of-Domain} Test. a) Heatmap showing the significance levels of $p$-values obtained by the Friedman Post-hoc Test and b) Critical difference (CD) plot showing the statistical significance of rankings achieved by the regression methods.\label{fig:stat-test-out-domain}}
			\end{figure}

			%
			%
			
			\subsection{Out-of-Sample Test}

			For the \textit{Out-of-Sample} testing, \textit{XGBoost} achieved the lowest error (median \textit{RMSE} score of 9.47) among the 11 regression methods. The three closest regression methods to \textit{XGBoost} are \textit{Random Forest} (median \textit{RMSE} of 9.67), \textit{Spline Continued Fraction} (median \textit{RMSE} of 10.99) and \textit{Gradient Boosting} (median \textit{RMSE} of 12.66). The \textit{Stochastic Gradient Descent}, without parameter estimation, performed the worst among all regression methods used in the experiment and due to the unreasonable high error observed in the runs we have omitted it from further analysis.

			\subsubsection{Statistical Significance Testing on the Results obtained for Out-of-Sample test\label{sec:res-stat-out-sample}}
			To evaluate the significance in results obtained by different regression methods for \textit{Out-of-Sample}, we applied a Friedman test for repeated measure~\cite{friedman1937use} for the 100 runs. Here, we computed the ranking of the methods for each of the runs based on the \textit{RMSE} score obtained in the test distribution of the \textit{Out-of-Sample} settings. It will help us determine if the experiment's techniques are consistent in terms of their performance. The statistical test found $p$-value = \num{1.9899e-183} which \textit{``rejected''} the \textit{null} hypothesis \textit{``all the algorithms perform the same''} and we proceeded with the posthoc test.

			We applied Friedman's posthoc test on the ranking of 10 regressors computed for the test \textit{RMSE} scores obtained for 100 runs of \textit{Out-of-Sample} test. In Fig.~\ref{fig:stat-test-out-sample} (a) the $p$-values obtained for the test are plotted as a heatmap. It is noticeable that there exist \textit{`no significant differences' (NS)} in performances of \textit{Spline Continued Fraction} (\texttt{Spnl-CFR}) with: \texttt{rf} and \texttt{grad-b}. 
			
			Additionally, we generated the Critical Difference (CD) diagram proposed in~\cite{demvsar2006statistical} to visualize the differences among the regressors for their median ranking. The CD plot used Nyemeni posthoc test and placed the regressors on the $x$-axis of their median ranking. It then computes the \textit{critical difference} of rankings between them and connects those which are closer than the critical difference with a horizontal line denoting them as statistically \textit{`non-significant'}.

			We plot the CD graph, in Fig.~\ref{fig:stat-test-out-sample} (b), using the implementation from Orange data mining toolbox~\cite{Python:Orange} in Python.  The Critical Difference (CD) is found to be $1.25$. We can see that the \texttt{xg-b} ranked \nth{1} among the regressors with \textit{`no significant difference'} with \nth{2} ranked \texttt{rf}. The median ranking of the proposed \textit{Spline Continued Fraction} is ranked \nth{3} with \textit{`no significant differences'} in the performance rankings of \texttt{rf} and \texttt{grad-b}.

			%
			%
			\subsection{Out-of-Domain Test}
			
			For the task of \textit{Out-of-Domain} prediction, the \textit{Spline Continued Fraction} regressor exhibited the best performance (median \textit{RMSE} score of 36.3) among all regression methods used in the experiment (in Table~\ref{tab:res-all-models}). Three closest regressors to the proposed \textit{Spline Continued Fraction} method are \textit{XGBoost} (median \textit{RMSE}=37.3), \textit{Random Forest} (median \textit{RMSE}=38.1) and \textit{Gradient Boosting} (median \textit{RMSE}=39.6).

			\subsubsection{Statistical Significance Testing on the Results obtained for Out-of-Domain test\label{sec:res-stat-out-domain}}
			To test the significance of the results obtained by different regression methods for \textit{Out-of-Domain} test, we employed the same statistical test used for \textit{Out-of-Sample} (in Sec.~\ref{sec:res-stat-out-sample}). The test returned a $p$-value = \num{1.2065e-156} which \textit{``rejected''} the \textit{null} hypothesis and we proceeded with the posthoc test.
			
			The $p$-values obtained for the posthoc test are plotted as a heatmap in Fig.~\ref{fig:stat-test-out-domain} (a) for \textit{Out-of-Domain} test. It is noticeable that there exist \textit{`no significant differences'} (NS) in performances of \textit{Spline Continued Fraction} (\texttt{Spnl-CFR}) with \textit{Random Forest} (\texttt{rf}) and \textit{XGBoost} (\texttt{xg-b}). There is also no significant difference in performance ranking of \textit{Linear Regression} (\texttt{l-regr}) with \texttt{mlp, l-svr, krnl-r} and \texttt{grad-b}.

			We plot the Critical Difference (CD) graph, in Fig.~\ref{fig:stat-test-out-domain} (b), for \textit{Out-of-Domain} test. The Critical Difference (CD) is $1.3898$. From the critical difference plot; it is evident that the top three methods in \textit{Out-of-Domain} prediction are \textit{Spline Continued Fraction}, \textit{XGBoost} and \textit{Random Forest}. We can see that the average ranking of \texttt{Spln-CFR} is very close to 2, which is the best-ranking performance among the 10 regressors. There is \textit{no significant difference}  of \textit{Spline Continued Fraction} with the \nth{2} best-ranked method, \textit{XGBoost} (\texttt{xg-b} with average ranking is between \nth{2} and \nth{3}), in \textit{Out-of-Domain} predictions.
			
			\subsubsection{Runtime Required by the methods for out-of-domain test}
			%
			%
			\begin{figure}
				\centering
				\includegraphics[width=10.5 cm]{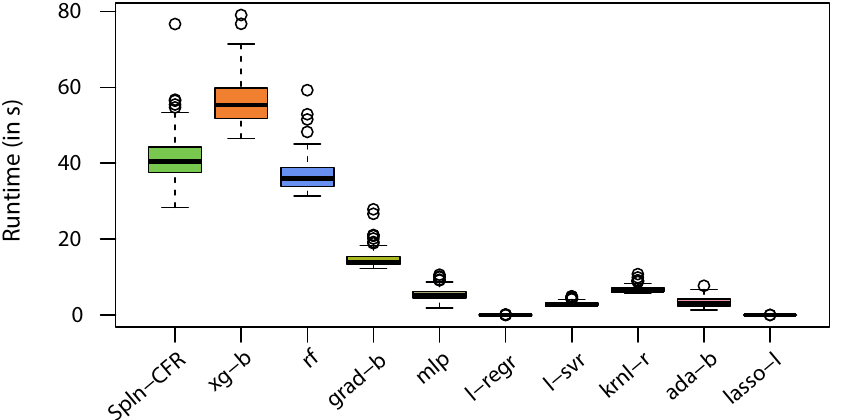}
				\caption{Run-time (in seconds) required for model building and predicting by the regressors for 100 runs of the \textit{Out-of-Domain} test, where samples with the lowest 90\% of critical temperatures were drawn to be the training data, with an equal number of samples constitute the test data (but these were withdrawn from the top 10\% highest critical temperatures).\label{fig:runtime-out-domain}}
			\end{figure}
			
			Fig.~\ref{fig:runtime-out-domain} shows the running time required by each of the regression methods (in s) for the 100 runs of \textit{Out-of-Domain} test. We can see that the \textit{Linear Regression} (\nth{50} percentile runtime of 0.02~s and maximum runtime 0.158~s) and Lasso lars (\nth{50} percentile 0.013~s and maximum of 0.027~s) required lowest running times. \textit{XGBoost} (\texttt{xg-b}) required the most amount of CPU time (\nth{50} percentile runtime of 55.33~s and maximum 79.05~s). On the other hand, \textit{Random Forest} and the proposed \textit{Spline Continued Fraction} Regression required nearly similar running time (\nth{50} percentile runtime of 36.88~s and 41.65~s for \texttt{rf} and \texttt{Spln-CFR}, respectively) for the \textit{Out-of-Domain} test.

			\section{Discussion}
			

			\begin{figure}[H]
				\centering
				\subfloat[\textit{Linear Regression}]{\includegraphics[width = 0.33\linewidth]{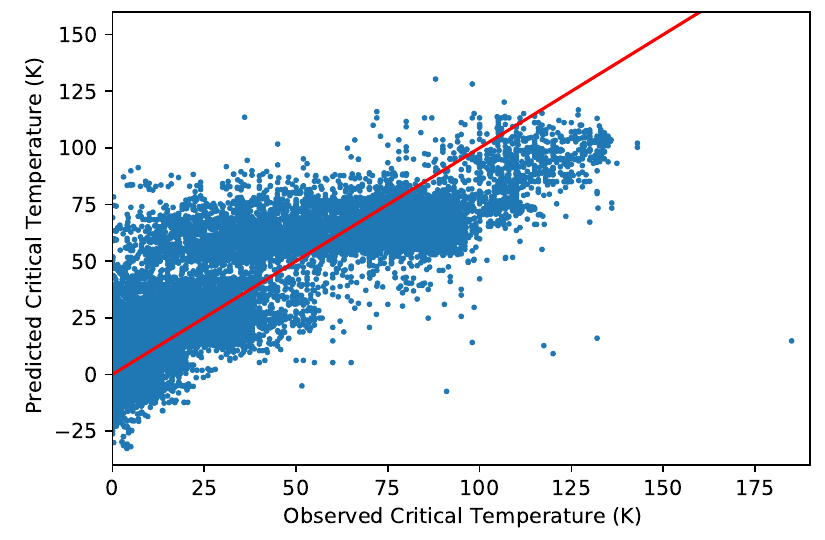}} \hfill 
				\subfloat[\textit{XGBoost}]{\includegraphics[width = 0.33\linewidth]{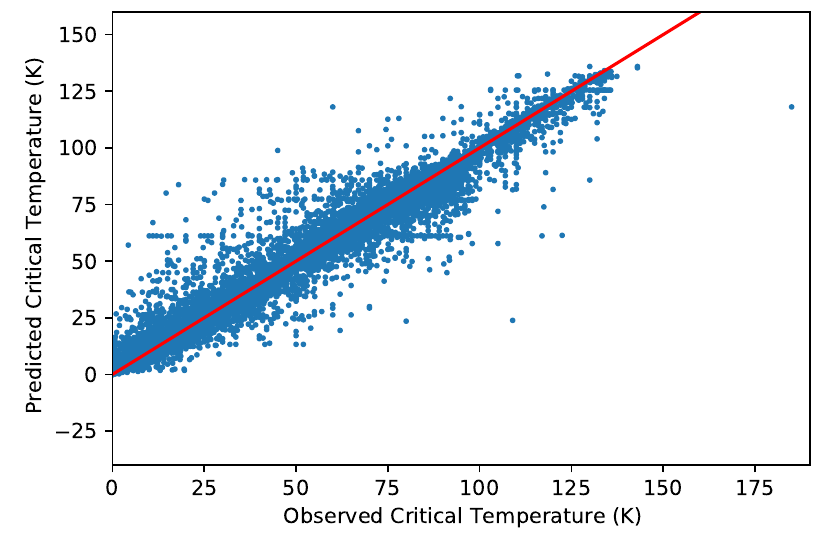}} \hfill
				\subfloat[\textit{Spline Continued Fraction}]{\includegraphics[width = 0.33\linewidth]{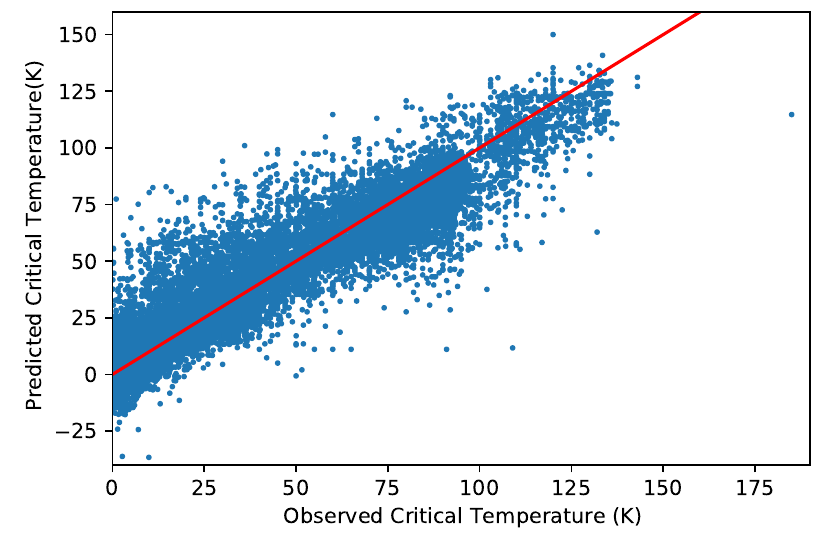}}
				\caption{\textit{Out-of-Sample} Test results showing Predicted vs actual temperatures for entire data with regression models trained on the training data. a) Results replicate \textit{Linear Regression} outcome from Hamidieh, b) \textit{XGBoost} and b) \textit{Spline Continued Fraction} model.\label{fig:out-sample}}
			\end{figure}
			
			To illustrate on the performance of models in the \textit{Out-of-Sample} study, we employed \textit{Linear regression}, \textit{XGBoost} and \textit{Spline Continued Fraction} on the training set and plotted the prediction vs actual temperatures for the entire dataset (in Fig.~\ref{fig:out-sample}). We show that we were able to reproduce the result of the \textit{Out-of-Sample} test from Hamidieh \cite{Superconductivity:2018-Hamidieh-DataDriven}
			Fig.~\ref{fig:out-sample} (a), 
			with \textit{RMSE} of 17.7. The \textit{Out-of-Sample} model for \textit{Spline Continued Fraction} and \textit{XGBoost} model are used to predict the critical temperature for the entire dataset. 
			Together, the figures show that \texttt{Spln-CFR} performed better in modelling \textit{Out-of-Sample} critical temperatures than that of \textit{Linear Regression}, particularly for larger temperatures.

			\begin{figure}[h]
				\centering
				\subfloat[\textit{Linear Regression}]{\includegraphics[width = 0.33\linewidth]{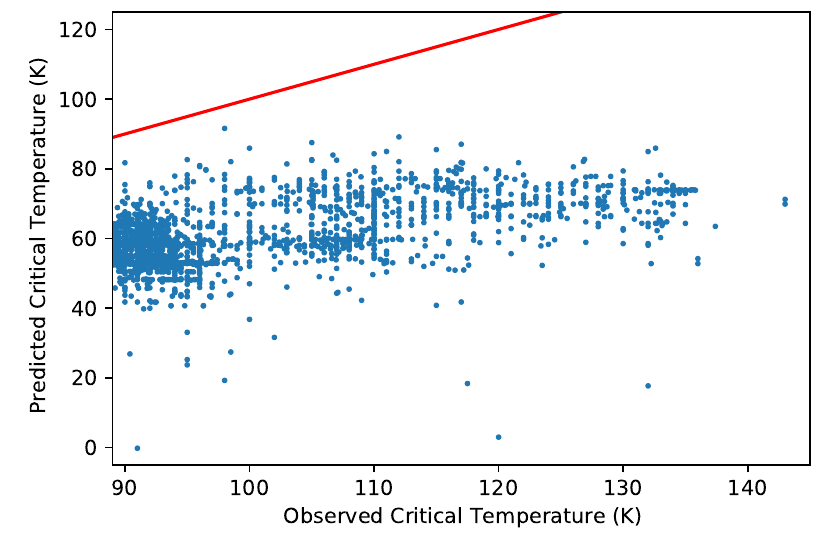}} \hfill 
				\subfloat[\textit{XGBoost}]{\includegraphics[width = 0.33\linewidth]{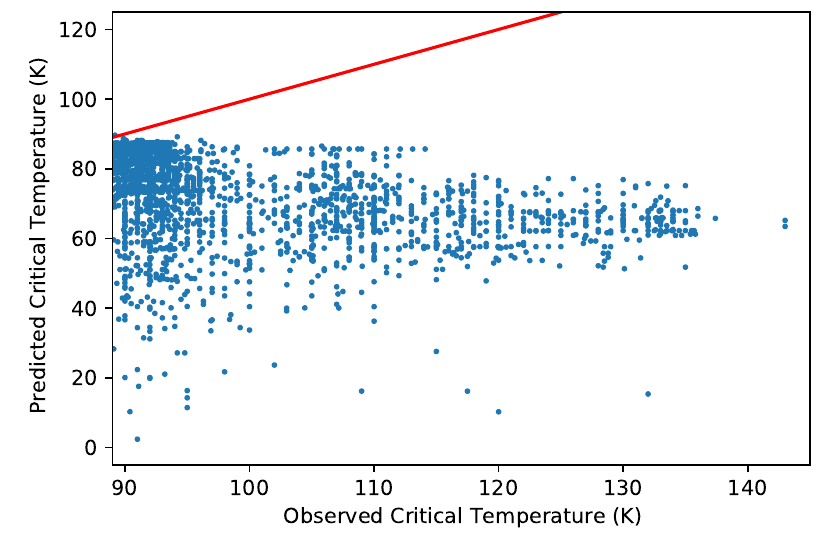}} \hfill
				\subfloat[\textit{Spline Continued Fraction}]{ \includegraphics[width = 0.33\linewidth]{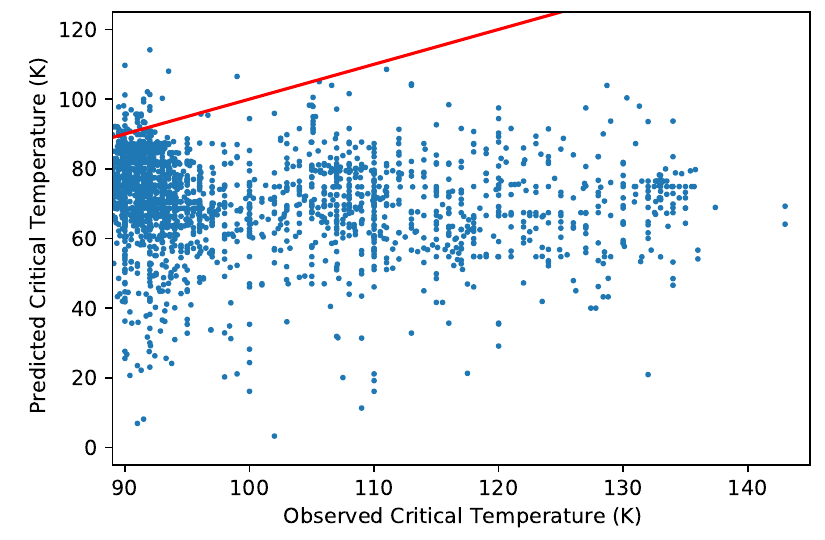}}
				\caption{\textit{Out-of-Domain} Test results showing Predicted vs actual temperatures of the samples for the highest 10\% critical temperatures, where a model is fitted using the samples with the lowest 90\% critical temperatures. We have shown the $x$-axis values up to 145~K which only left an extreme value (185~K) out of the visual area.  Results of \textit{Out-of-Domain} test for a) \textit{Linear Regression} with \textit{RMSE} of  41.3 b) \textit{XGBoost} with \textit{RMSE} of 36.3 and c) \textit{Spline Continued Fraction} model's with \textit{RMSE} of 34.8.\label{fig:out-domain}}
			\end{figure}
			
			Fig.~\ref{fig:out-domain} shows actual vs predicted critical temperature for the \textit{Out-of-Domain} test for \textit{Linear Regression}, \textit{XGBoost} and \textit{Spline Continued Fraction} models. We recall that in \textit{Out-of-Domain} settings, we trained each of the models with the samples from the bottom 90\% of the observed temperature (which is < 89~K). We measured the samples' testing performance with the top 10\% of the observed critical temperatures (containing 2126 samples in the test set).

			\begin{table}[H]
				\caption{Predicted vs. Actual critical temperatures for the materials with the top 20 predicted temperatures in the {Out-of-Domain} study,	i.e.\ the one in which the lowest 90\% of critical temperature samples were used for drawing the training data. The average values of the critical temperatures ($\bar{x}$), the average relative error ($\bar{\eta}$), and the Root mean squared error (\textit{RMSE} as \textit{rm}) of these materials for the top 20 predictions (which are not necessarily the same since they depend on the models) are shown in the last rows.
					\label{tab:out-domain-top-20-temp}}
					\begin{tabularx}{\textwidth}{*{21}{R}}
						\toprule
						{} & \multicolumn{2}{r}{\textbf{Spln-CFR}} & \multicolumn{2}{r}{\textbf{xg-b}} & \multicolumn{2}{r}{\textbf{rf}} & \multicolumn{2}{r}{\textbf{grad-b}} & \multicolumn{2}{r}{\textbf{mlp}} &  \multicolumn{2}{r}{\textbf{l-regr}} & \multicolumn{2}{r}\textbf{{l-svr}} & \multicolumn{2}{r}{\textbf{krnl-r}} &  \multicolumn{2}{r}{\textbf{ada-b}} & \multicolumn{2}{r}{\textbf{lasso-l}} \\
						\cline{2-21}
						{} & {y} & pred & y & pred & y & pred & y & pred & y & pred & y & pred & y & pred & y & pred & y & pred & y & pred\\
						\midrule
						{} &     {92.00} &         114.14 &   89.20 &      89.64 & 91.19 &    87.89 &     89.50 &        83.44 & 109.00 &    100.81 &     98.00 &        91.59 &   112.00 &       94.81 &     98.00 &        91.02 &    89.50 &       58.63 &      89.00 &         27.06 \\
						{} &      {90.00} &         109.69 &   94.20 &      89.19 & 89.90 &    87.88 &     89.90 &        83.44 & 124.90 &    100.31 &    112.00 &        89.14 &   100.00 &       93.49 &    112.00 &        88.67 &    89.50 &       58.63 &      89.00 &         27.06 \\
						{} &     {111.00} &         108.54 &   89.88 &      88.69 & 90.00 &    87.88 &     90.50 &        83.44 & 114.00 &     99.70 &    105.00 &        87.53 &   132.60 &       93.49 &    105.00 &        86.84 &    89.70 &       58.63 &      89.00 &         27.06 \\
						{} &      {93.50} &         108.01 &   89.93 &      88.34 & 90.20 &    87.88 &     91.50 &        83.44 & 128.40 &     99.59 &    117.00 &        87.06 &   105.00 &       92.94 &    117.00 &        86.65 &    89.80 &       58.63 &      89.00 &         27.06 \\
						{} &     {99.00} &         106.50 &   90.00 &      88.15 & 90.90 &    87.88 &     90.00 &        83.42 & 127.40 &     99.53 &    100.00 &        85.92 &   115.00 &       92.93 &    100.00 &        85.88 &    89.80 &       58.63 &      89.00 &         27.06 \\
						{} &    {105.60} &         105.01 &   90.10 &      88.15 & 91.00 &    87.88 &     91.80 &        83.42 & 127.80 &     99.53 &    132.60 &        85.92 &   111.00 &       92.90 &    132.60 &        85.88 &    89.90 &       58.63 &      89.00 &         27.06 \\
						{} &      {113.00} &         104.35 &   91.00 &      88.15 & 92.00 &    87.88 &     90.00 &        82.22 & 130.10 &     98.76 &    115.00 &        85.50 &   110.00 &       92.84 &    115.00 &        85.51 &    90.00 &       58.63 &      89.00 &         27.06 \\
						{} &    {113.00} &         103.95 &   91.30 &      88.15 & 92.20 &    87.88 &     89.50 &        79.29 & 128.50 &     98.55 &    111.00 &        84.97 &   106.70 &       92.54 &    111.00 &        84.46 &    90.00 &       58.63 &      89.00 &         27.06 \\
						{} &      {106.60} &         103.95 &   96.10 &      88.15 & 92.40 &    87.88 &     90.00 &        79.29 & 128.40 &     98.45 &    132.00 &        84.96 &   126.90 &       91.73 &    132.00 &        84.42 &    90.50 &       58.63 &      89.00 &         27.06 \\
						{} &     {128.70} &         103.92 &   90.00 &      88.10 & 92.50 &    87.88 &     91.00 &        79.29 & 128.80 &     98.45 &    110.00 &        84.31 &   117.00 &       91.73 &    110.00 &        84.38 &    91.50 &       58.63 &      89.00 &         27.06 \\
						{} &      {91.80} &         102.10 &   91.40 &      88.10 & 92.74 &    87.88 &     91.80 &        79.29 & 131.40 &     98.33 &    106.70 &        83.95 &   126.80 &       91.30 &    106.70 &        82.97 &   100.00 &       58.63 &      89.00 &         27.06 \\
						{} &     {108.00} &         101.56 &   92.60 &      87.82 & 92.80 &    87.88 &     92.30 &        79.29 & 128.80 &     98.10 &    126.90 &        82.72 &   115.00 &       90.84 &     95.00 &        82.64 &   108.00 &       58.63 &      89.00 &         27.06 \\
						{} &      {92.00} &         101.32 &   91.60 &      87.53 & 93.00 &    87.88 &     90.00 &        78.85 & 128.70 &     93.96 &    105.00 &        82.63 &    95.00 &       90.80 &    105.00 &        82.01 &   110.00 &       58.63 &      89.00 &         27.06 \\
						{} &       {90.00} &         101.19 &   93.00 &      87.53 & 93.00 &    87.88 &     91.60 &        78.85 & 130.30 &     93.94 &     95.00 &        82.62 &   121.60 &       90.80 &    107.00 &        81.88 &   110.90 &       58.63 &      89.00 &         27.06 \\
						{} &      {105.10} &         100.50 &   93.80 &      87.49 & 93.05 &    87.88 &     89.10 &        78.79 & 131.30 &     93.93 &    107.00 &        82.47 &   100.00 &       90.78 &    126.90 &        81.82 &   114.00 &       58.63 &      89.00 &         27.06 \\
						{} &      {130.30} &         100.35 &   89.90 &      87.48 & 93.20 &    87.88 &     89.20 &        78.79 & 122.00 &     91.96 &    105.00 &        82.41 &   107.00 &       90.78 &    105.00 &        81.51 &   114.00 &       58.63 &      89.00 &         27.06 \\
						{} &      {93.00} &         100.24 &   90.00 &      87.48 & 93.40 &    87.88 &     89.40 &        78.79 & 123.50 &     91.64 &    126.80 &        82.12 &    90.00 &       90.63 &     90.00 &        81.40 &   116.00 &       58.63 &      89.10 &         27.06 \\
						{} &      {91.50} &         100.00 &   90.20 &      87.48 & 93.50 &    87.88 &     89.40 &        78.79 & 121.00 &     90.69 &     98.50 &        82.03 &    96.00 &       90.49 &    126.80 &        81.24 &   122.50 &       58.63 &      89.10 &         27.06 \\
						{} &       {91.50} &          99.18 &   90.90 &      87.48 & 91.80 &    87.75 &     89.40 &        78.79 & 115.00 &     90.14 &    112.00 &        82.03 &   128.70 &       90.48 &    117.00 &        80.89 &   127.00 &       58.63 &      89.10 &         27.06 \\
						{} &     {116.00} &          98.39 &   91.00 &      87.48 & 92.10 &    87.69 &     89.50 &        78.79 & 110.00 &     90.01 &    117.00 &        81.83 &   130.30 &       90.26 &    121.60 &        80.87 &   130.90 &       58.63 &      89.10 &         27.06 \\
						\midrule
						{$\bar{x}$:} & {103.08} & 103.64 & 91.31 & 88.03 & 92.044& 87.86 & 90.27 & 80.49 & 124.47 & 96.32 & 111.63 & 84.59 & 112.33 & 91.83 & 111.68 & 84.05 & 102.68 & 58.63 & 89.02 & 27.06\\
						
						{$\bar{\eta}$:} & \multicolumn{2}{r}{0.1085} &	\multicolumn{2}{r}{0.036} &	\multicolumn{2}{r}{0.0453} &	\multicolumn{2}{r}{0.1083} &	\multicolumn{2}{r}{0.224} &	\multicolumn{2}{r}{0.2351} &	\multicolumn{2}{r}{0.1733} &	\multicolumn{2}{r}{0.2389} &	\multicolumn{2}{r}{0.4187} &	\multicolumn{2}{r}{0.696}
						\\
						\textit{rm:} & \multicolumn{2}{r}{13.6023}	& \multicolumn{2}{r}{3.7753}	& \multicolumn{2}{r}{4.3261}	& \multicolumn{2}{r}{10.0078}	& \multicolumn{2}{r}{28.9783}	& \multicolumn{2}{r}{29.3265}	& \multicolumn{2}{r}{23.9282}	& \multicolumn{2}{r}{30.2426}	& \multicolumn{2}{r}{46.2473}	& \multicolumn{2}{r}{61.96}
						\\
						\bottomrule
					\end{tabularx}
			\end{table}
			
			Another set of observed results are interesting for discussion and might be relevant for future research directions.
			In Table~\ref{tab:out-domain-top-20-temp}, we report the top 20 predicted vs, actual ($y$) temperatures for all ten regression methods for \textit{Out-of-Domain} test of a single run. The last row of the table shows the average of the corresponding (actual) critical temperature for the materials with the highest 20 predicted values by each of the models. Interestingly, \textit{XGBoost}'s top 20 predictions of the critical temperatures are all below 90~K (in the range of 87.48 to 89.64~K). Similarly, \textit{Random Forest}'s top 20 predictions are in the range of 87.69 to 87.89~K. The top 20 predicted critical temperatures by the \textit{Linear Regression} are in the range of 81.83 to 91.59~K. In contrast, the top 20 predicted critical temperature by \textit{Spline Continued Fraction} varies from 98.39 to 114.14~K, which, in comparison, has the highest starting and ending values among all regressors. We also reported the average temperature ($\bar{x}$), average relative errors ($\bar{\eta}$) and \textit{RMSE} score computed for the top 20 predictions. \textit{XGBoost} has showed the lowest value for both $\bar{\eta}$ (0.036) and \textit{RMSE} ($3.775$) among 10 regressors. In terms of those scores, the proposed \texttt{Spln-CFR} is \nth{4} position. However, if we look at the average of predictions,  \texttt{Spln-CFR} has the highest average prediction temperatures for the top 20 predictions in \textit{Out-of-Domain} tests.

			Since all the actual critical temperatures of the test set in \textit{Out-of-Domain} settings are $\geq 89$~K, it is relevant to evaluate for how many of these samples each regression method was able to predict above that value. Here, we considered the predicted value as \textbf{P}  = \textit{critical temperature value} $\geq 89$~K (denoted as `P', for positive) and \textbf{N} = \textit{critical temperature value} < 89~K (denoted as `N', for negative). In Table~\ref{tab:res-sum_out_domain}, we reported the number of samples for which each of the methods predicted a temperature value in the P and N category for the whole testing set of \textit{Out-of-Domain} test. It is found that only six regression methods predicted the critical temperature being $\geq 89$~K for at least one sample. Both \textit{Linear Regression} and \textit{XGBoost} predicted two sample's temperatures with critical temperature $\geq 89$~K. Kernel Ridge predicted only one sample's value within that range. MLP Regressor and Linear SVR predicted it for 21 and 34 samples, respectively. The proposed \textit{Spline Continued Fraction} predicted 108 sample's value $\geq 89$~K, which is the best among all regression methods used in the experiments.

			\begin{table}[H] 
				\caption{Number of times the methods predicted a critical temperature value $T_c \geq 89$~K (denoted as `P', for positive) and $T_c < 89$~K (denoted as `N' for Negative) for {Out-of-Domain} test.\label{tab:res-sum_out_domain}}
				\begin{tabularx}{\textwidth}{CCC}
					\toprule
					\multirow{2}{*}{\textbf{Regressor}} & \multicolumn{2}{r}{\textbf{Out-of-domain predicted critical temperature,} $T_c$}\\
					\cline{2-3}
					{} &  {P ($T_c$ $\geq 89$~K)} & {N ($T_c$ < 89~K)} \\
					\midrule
					Spln-CFR &                       108 &                      2018 \\
					xg-b &                         2 &                      2124 \\
					rf &                         0 &                      2126 \\
					grad-b &                         0 &                      2126 \\
					mlp &                        21 &                      2105 \\
					l-regr &                         2 &                      2124 \\
					l-svr &                        34 &                      2092 \\
					krnl-r &                         1 &                      2125 \\
					ada-b &                         0 &                      2126 \\
					lasso-l &                         0 &                      2126 \\
					\bottomrule
				\end{tabularx}
			\end{table}

			\begin{table}[H]
				\centering
				\caption{Inter-rater agreement between the pairs of regressor methods where the resulting models were able to predict at least one positive temperature value ($T_c \geq 89$~K).\label{tab:kappa-out-domain}}
				\begin{tabularx}{\textwidth}{CCRC}	
					\toprule
					\textbf{Rater 1} & \textbf{Rater 2} &  \textbf{Value of Kappa ($\kappa$)} &   \textbf{Level of Agreement} \\
					\midrule
					Spln-CFR &    xg-b & -0.001851 &    No Agreement \\
					Spln-CFR &     mlp &  0.030476 &  None to Slight \\
					Spln-CFR &  l-regr &  0.016365 &  None to Slight \\
					Spln-CFR &   l-svr &  0.104988 &  None to Slight \\
					Spln-CFR &  krnl-r & -0.000933 &    No Agreement \\
					xg-b &     mlp & -0.001721 &    No Agreement \\
					xg-b &  l-regr & -0.000942 &    No Agreement \\
					xg-b &   l-svr & -0.001780 &    No Agreement \\
					xg-b &  krnl-r & -0.000628 &    No Agreement \\
					mlp &  l-regr & -0.001721 &    No Agreement \\
					mlp &   l-svr &  0.208516 &            Fair \\
					mlp &  krnl-r & -0.000899 &    No Agreement \\
					l-regr &   l-svr &  0.053874 &  None to Slight \\
					l-regr &  krnl-r &  0.666457 &     Substantial \\
					l-svr &  krnl-r & -0.000915 &    No Agreement \\
					\bottomrule
				\end{tabularx}
			\end{table}
			
			We look at the consensus between regression methods in \textit{Out-of-Domain} prediction. Only five regressors (\texttt{Spln-CFR}, \texttt{xg-b}, \texttt{mlp}, \texttt{l-regr}, \texttt{l-svr} and \texttt{krnl-r}) which were able to predict at least one positive value (critical temperature $\geq 89$~K). We computed pairwise inter-rater agreement statistics, Cohen's kappa~\cite{cohen1960coefficient}, for those regression methods. We tabulated the value of Kappa($\kappa$) ordered by highest to lowest and outlined the level of agreement in Table~\ref{tab:kappa-out-domain}. We can see that in most of the cases, there is either \textit{``No''} (9 cases) or \textit{``None to Slight''} (4 cases) agreement exists between the pairs of regressors. We witness such behaviour in the agreement between the pairs formed with \texttt{Spln-CFR} and each of the other five methods. MLP Regressor and Linear SVR have \textit{``Fair''} agreement in the predictions. We witnessed the highest value $\kappa=0.67$ for \textit{Linear Regression} and Kernel Ridge, which yields a \textit{``Substantial''} agreement.
			
			\subsection{Extrapolation Capability of the Regressors in General}

			As all of the results presented in this work are for a special case of finding models for the 
			extrapolation of the critical temperature of superconductors, we included more robust experimental outcomes with a set of six datasets used in \cite{DBLP:conf/cec/SunM19}. This additional test will help us to evaluate the extrapolation capabilities of the regressors in other problem domains.
			
			Jerome Friedman proposed a Multivariate Adaptive Regression Splines (MARS) algorithm in~\cite{friedman1991MARS} 
			which aggregates multiple linear regression models throughout the range of target values. We used the implementation of the MARS algorithm from py-earth Python library~\footnote{\url{https://contrib.scikit-learn.org/py-earth/content.html\#multivariate-adaptive-regression-splines}}. We included a comparison of MARS with \texttt{Spln-CFR} and other regressors for extrapolation capability. 
			
			Here, the samples from each of the datasets were sorted based on the target value. Then we split it into the out-of-domain setting by taking samples with lower 90\% target values as train and higher 10\% target values as a test. We uniformly at random took half of the samples from the out-of-domain train to build the model and the same ratio from the out-of-domain test sets for prediction for each of the 100 independent runs. We applied min-max normalization on the train set and used the same distribution to normalize the test set.

			\begin{table}[H]
				\centering
				\caption{Number of times the prediction by a method in the range of out-of-domain threshold in test sets for 100 repeated runs on six datasets form~\cite{DBLP:conf/cec/SunM19}\label{tab:cec-od-range}}
				\begin{tabularx}{\textwidth}{CRCRCR}
					\toprule
					\textbf{Regressor} & \textbf{in Range} & \textbf{Regressor} & \textbf{in Range} & \textbf{Regressor} & \textbf{in Range} \\
					\midrule
					Spln-CFR  &  13560   & grad-b    & 1227     & ada-b     & 0       \\
					MARS      & 3716     & mlp       & 1158      & lasso-l   & 0       \\
					l-regr    & 2594     & xg-b      & 826      & rf        & 0       \\
					l-svr     & 2045     & krnl-r    & 735      & sgd-r     & 0   \\
					\bottomrule
				\end{tabularx}
			\end{table}

			\begin{figure}[H]
				\centering
				\includegraphics[width =10.5 cm]{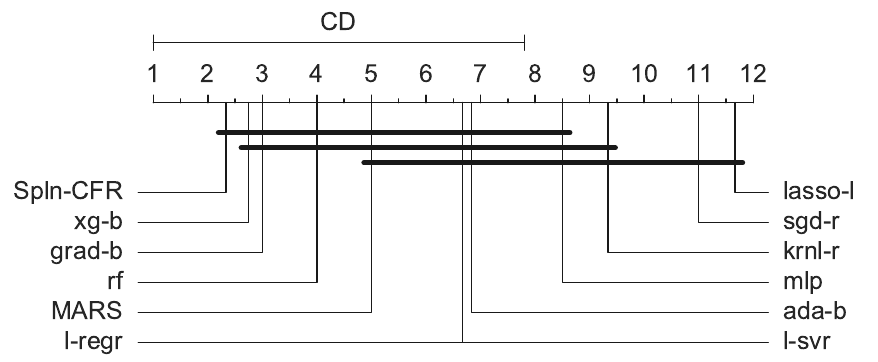}
				\caption{Critical difference (CD) plot showing the statistical significance of rankings achieved by the regression methods for 100 runs on the six datasets form~\cite{DBLP:conf/cec/SunM19}.
					\label{fig:cd-100runs-cec2019}}
			\end{figure}
			
			We have analyzed their performance statistically (in Fig.~\ref{fig:cd-100runs-cec2019}) and found that MARS has a median ranking of 5, and is statistically significantly different from only \texttt{krnl-r, sgd-r} and \texttt{lasso-l}. However, the proposed \texttt{Spln-CFR} has achieved the first rank among all the methods with a median ranking between two to three. The predictions by each model are de-normalized to count the number of predictions above the threshold (the maximum target value in the training portion of data) in out-of-domain settings. We show the complete outcome in Table~\ref{tab:cec-od-range}. These counts show that the \texttt{Spln-CFR} has the highest number of predictions (13560) followed by MARS (3716) and \texttt{l-regr} (2594) are in the range. These results demonstrate the strength of the regressors for their extrapolation capability.

			\section{Conclusions}
			
			We give a brief summary of some of the results observed on this new technique:
			
			\begin{itemize}
				\item For the \textit{Out-of-Sample} study, the median \textit{RMSE} obtained for 100 independent runs, the proposed \texttt{Spln-CFR} is in the top three methods (in Table~\ref{tab:res-all-models}).
				\item For the statistical test of \textit{Out-of-Sample} rankings, \texttt{Spln-CFR} is statistically similar to the \nth{2} ranked method (\textit{Random Forest}) in Fig.~\ref{fig:stat-test-out-sample} (b).
				\item For \textit{Out-of-Domain} median \textit{RMSE} obtained for 100 runs, the proposed \texttt{Spln-CFR} is the top method (ranked \nth{1} in Table~\ref{tab:res-all-models}).
				\item For the statistical test of \textit{Out-of-Domain} rankings, in Fig.~\ref{fig:stat-test-out-domain} (b), \texttt{Spln-CFR} is the best method (median ranking is close to 2) and statistically similar to the second best regressor, \textit{XGBoost} (with a median ranking between 2 and 3).
				\item \texttt{Spln-CFR} correctly predicted that 108 unique materials have critical temperature values that are greater than or equal to 89~K in  \textit{Out-of-Domain} test (close to twice the number of all other regression methods tested combined which was 60) (Table~\ref{tab:res-sum_out_domain}).
			\end{itemize}
			
			Table \ref{tab:out-domain-top-20-temp} also reveals interesting characteristics of all methods that deserve further consideration as an area of research. 
			First, note that the 20 top materials for each of the methods are not necessarily the same, although some intersections obviously may exist. 
			In the \textit{Out-of-Domain} study, the top 20 predicted critical temperature values by \texttt{Spln-CFR} were all above 98.9~K (with 18 being above 100~K). The average \textit{RMSE} critical temperature on this set (103.64 K) is nearly the same as the one predicted (103.08~K). The \textit{RMSE} of \texttt{xg-b}, however, is nearly three times smaller, but the method's top predictions are materials with relatively smaller values (average of 91.31~K). We observed, for the collected information of materials in the dataset, the top suggestions of critical temperatures in superconductors are closer to the measured temperature, at least on the average, by the \texttt{Spln-CFR}. Therefore, the usage of \texttt{Spln-CFR} as a surrogate model to explore the possibility of testing the superconductivity in materials may bring better returns.

			Interestingly, we have also observed a similar behavior of \texttt{xg-b} with other multivariate regression techniques, but also important differences worth noting. For instance, \textit{Linear Regression}, perhaps the simplest scheme of them all, has an interesting behavior: the top 20 highest predictions are all in the range $[81.83, 91.59]$~K while the actual values are in the interval $[98.00, 132.60]$~K. For the multi-layer perceptron method (\texttt{mlp}), the top 20 highest predictions are all in the range $[90.01, 100.81]$~K, yet true values are in the interval $[109.00, 131.40]$~K. This means that trained using the MSE, these techniques could still give valuable information about materials that could be prioritized for testing if we better consider the ranking given to several materials and have less concern about the predicted value.

			Overall, the results show the limitations of the current dataset. One possible limitation is the lack of other useful molecular descriptors that can bring important problem-domain knowledge about the structure of the materials and their properties. In addition, it is also possible that a careful ``segmentation'' of the different materials is necessary. In some sense, the results of the experiments presented here may help the AI community reflect on how to do these analyses and motivate a closer collaboration with superconductivity specialists to provide other molecular descriptors.
			
			We actually often compare the inherent difficulties in prediction in this dataset to other areas on which some of us have been working extensively (like the prediction of survivability in breast cancer using transcriptomic data). In both cases, without separating the training samples into meaningful subgroups, the models obtained generalised poorly.	This said, one of the reasons that our continued fraction-based method may be doing just a bit better in the generalisation test in our \textit{Out-of-Domain} study, is that there might be some structural similarities in the set of compounds used to define the continued fraction approximation at the highest temperatures in the training set, then, indirectly perhaps, from the molecular descriptors present in these samples some useful information exists which the continued fraction representation has exploited. We will investigate this hypothesis in a future publication where we aim to include more relevant problem-domain information, in collaboration with specialists, to benefit from the structure and known properties of the actual compounds.
			
			In terms of future research on the algorithm we propose here, it is clear that \texttt{Spln-CFR} is already a promising approach that has some obvious extensions worth considering in the future, for instance, the inclusion of \textit{bagging} and \textit{boosting} techniques which can improve the \textit{Out-of-Sample} performance. In addition, we consider that learning with modifications of the MSE in the training set may lead to better performance for the \textit{Out-of-Domain} scenario, and we plan to conduct further research in that area as well. 
			
			\section*{CRediT author statement}
			{Conceptualization, P.M.; methodology, P.M., K.H.; software, M.N.H, K.H., J.S. and J.C.O; validation, M.N.H., K.H. and P.M.; formal analysis, M.N.H, P.M.; investigation, P.M., M.N.H, K.H.; data curation, K.H., J.S., J.C.O.; visualization, K.H., M.N.H.; supervision, P.M.; project administration, P.M., M.N.H.; funding acquisition, P.M. All authors have participated in writing, read and agreed to the published version of the manuscript.}
			
			\section*{Acknowledgements}
			{This work was supported by the Australian Government through the Australian Research Council's Discovery Projects funding scheme (project DP200102364). P.M. acknowledges a generous donation from the Maitland Cancer Appeal. This work has been supported by the University of Newcastle and Caltech Summer Undergraduate Research Fellowships (SURF) program. In particular, SURF Fellows J. Sloan and K. Huang acknowledge the gifts from Samuel P. and Frances Krown, and Arthur R. Adams, respectively, for their generous donor support to their activities through the SURF program.}

			\bibliographystyle{unsrtnat}
			\bibliography{references}  

				
				
				
				

		\end{document}